\UseRawInputEncoding
\documentclass{article}
\usepackage{PRIMEarxiv}

\usepackage[utf8]{inputenc} 
\usepackage[T1]{fontenc}    
\usepackage{hyperref}       
\usepackage{url}            
\usepackage{booktabs}       
\usepackage{amsfonts}       
\usepackage{nicefrac}       
\usepackage{microtype}      
\usepackage{lipsum}
\usepackage{fancyhdr}       
\usepackage{graphicx}       
\usepackage{subfigure}
\graphicspath{{media/}}     
\usepackage{multirow}
\usepackage{diagbox}
\title{Chinese Grammatical Error Correction Based On Knowledge Distillation
}

\author{
  Peng Xia, Yuechi Zhou, Ziyan Zhang, Zecheng Tang \\
  School of Computer Science and Technology\\
  Soochow University \\
  \texttt{\{1909401089,1924401017,1928401062,zctang\}@stu.suda.edu.cn} \\
   \And
  Juntao Li \\
  Institute of Artificial Intelligence\\
  Soochow University \\
  \texttt{ljt@suda.edu.cn} \\
}

\begin{document}
\maketitle

\begin{abstract}
In view of the poor robustness of existing Chinese grammatical error correction models on attack test sets and large model parameters, this paper uses the method of knowledge distillation to compress model parameters and improve the anti-attack ability of the model. In terms of data, the attack test set is constructed by integrating the disturbance into the standard evaluation data set, and the model robustness is evaluated by the attack test set. The experimental results show that the distilled small model can ensure the performance and improve the training speed under the condition of reducing the number of model parameters, and achieve the optimal effect on the attack test set, and the robustness is significantly improved. Code is available at \url{https://github.com/Richard88888/KD-CGEC}.
\end{abstract}


\section{Introduction}
Grammatical Error Correction (GEC) is an important task in the field of natural language processing. It aims to detect whether there are Grammatical errors in sentences and automatically correct the detected grammatical errors. Grammatical error correction has important applications in writing assistants \cite{ghufron2018role,napoles2017jfleg}, search engines \cite{martins2004spelling,gao2010large}, speech recognition \cite{karat1999patterns,wang2020asr}, etc. 

At present, the research on grammatical error correction is mainly focused on English and Chinese. The study of English grammatical error correction started relatively early, in the 1980s when the study related to grammatical error correction has appeared. Nowadays, there are many methods for English grammatical error correction. Compared with English grammatical error correction, the research on Chinese grammatical error correction started relatively late, and the Chinese grammar rules are complex, which determines that the complexity of Chinese grammatical error correction is higher than that of English.

In the early stage, most Chinese grammatical error correction methods were based on rules and statistics. In 2009, Wu et al. \cite{wu2009sentence} proposed to supplement traditional language modeling by using relative location language model and parsing template language model. However, this method requires a lot of manpower to design appropriate rules, and the customized rules are difficult to popularize, so there are great limitations. To this end, researchers proposed the method of constructing classifiers for different errors and combining these classifiers into a hybrid system \cite{rozovskaya2014illinois}. In 2012, Yu and Chen \cite{yu2012detecting} proposed a classifier method to detect Chinese sentence word order errors, and integrated grammatical features, network corpus features and disturbance features, which is of great help to detect Chinese word order errors.

Then, some researchers began to regard syntax correction as a Sequence-to-Sequence (\textit{Seq2Seq}) task, and proposed a solution based on Statistical Machine Translation (SMT) model \cite{brockett2006correcting}, which have been achieved good results \cite{chollampatt2017connecting}. In 2014, Zhao et al. \cite{zhao2015improving} proposed a statistical machine translation model based on grammar and hierarchical phrase, and explored the effect of corpus expansion. However, the generalization ability of statistical methods is poor, and it is difficult to obtain distant context information, so the performance is limited. 

In recent years, with the development of deep learning technology, Neural Machine Translation (NMT) model has gradually become the preferred model for grammar error correction. In 2016, Yuan et al. \cite{yuan2016grammatical} first applied neural machine translation to the task of grammatical error correction. This method generated grammatically correct sentences by inputting grammatically incorrect sentences into a model composed of bidirectional recurrent neural network, and achieved good results. In order to enhance the focus of the model on local errors in spelling and deformation, Ji \cite{ji2017nested} proposed a hybrid machine translation model based on nested attention layer in 2017, which combines word-level and character-level information. In 2018, Chollampatt et al. \cite{chollampatt2018multilayer} proposed the method of multi-layer convolutional neural network to better capture local context information, thus improving the coverage of syntax error correction. Moreover, through model integration and combination of N-gram language model and editing features, the method based on statistical machine translation was superior to the method based on statistical machine translation for the first time in terms of syntax and fluency. In 2020, Wang et al. \cite{wang2020jiyu} proposed a Chinese syntax error correction model of Transformer enhanced architecture based on multi-head self-attention mechanism, which used dynamic residual structure and combined the output of different neural modules to enhance the ability of the model to capture semantic information. In 2020, Zhang et al. \cite{zhang2020mian} proposed personalized grammar correction for the first time, aiming at the grammatical errors made by people with different levels of Chinese learning, and corrected the errors of different feature types respectively, thus significantly improving the performance of the model. In 2020, Zhao et al. \cite{zhao2020maskgec} proposed a dynamic random masked method, which enabled the original one-way syntax correction model to learn the bidirectional language representation and achieved the best performance at that time. 

Due to the lack of large-scale Chinese grammar correction of high quality parallel corpus, various researchers began to pay attention to the size of the corpus collection and error range of coverage, in 2018 Xie et al \cite{xie2018noising} used back translation methods synthesis of pseudo corpus, then pseudo corpora into training in order to enhance the data size, at the same time, in order to guarantee the wrong type of coverage, Xie et al. \cite{xie2018noising} produced diversified error types by perturbing the decoding process. In 2020, Wang et al. \cite{wang2020jiyu} proposed a rule-based corrupt corpus set method to increase error coverage. In 2021, Tang et al. \cite{tang2021ji} proposed a data augmentation method combining word and character granularity noise to simultaneously expand the data scale and error coverage, which is referred to as Trans-fuse in the following. 

The above methods are generally based on the model architecture of \textit{Seq2Seq}. Although good results can be obtained, the number of parameters is usually large. For example, the number of parameters in Transformer-big \cite{vaswani2017attention} is 357,831,680. In addition, we further find that the robustness of existing syntax error correction models is poor. To address this issue, we proposed a process called Knowledge Distillation \cite{hinton2015distilling} to perform model compression while improving model robustness. 

Knowledge distillation (KD) is a common approach in machine learning that uses simpler models to model the predictions of more complex models, with the aim of reducing the computational cost and memory requirements of deployment. It has been successfully applied in many fields of deep learning, such as object detection \cite{chen2017learning} and natural language processing \cite{cui2017knowledge, yu2017visual}. Fathullah et al. \cite{fathullah2021ensemble} first applied Ensemble Distillation and Ensemble Distribution Distillation to sequential tasks. The application of these two distillation methods in English grammar error correction is studied. To the best of our knowledge, no researchers have applied knowledge distillation to the task of Chinese grammar correction.

In this paper, firstly a Chinese grammatical error correction model based on Transformer architecture is trained by the data augmentation method combining word and character granularity noise. Secondly, the method of knowledge distillation is used to reduce the parameters of the large model and accelerate the convergence speed in the training stage. Finally, the experimental results show that the method of knowledge distillation can improve the performance and robustness of the grammar error correction model while reducing the number of parameters.

\section{Related Work \& Methodology}
\label{sec:headings}

\subsection{Transformer Model}
Transformer model \cite{vaswani2017attention} is based on the encoder-decoder architecture. The encoder is responsible for encoding the input sentence into a high-dimensional hidden state vector, and the decoder decodes the implicit semantic vector and outputs the prediction result at the current moment based on the output result at the last moment.

The encoder is composed of $N$ neural network blocks, and each neural network block contains a multi-head attention sublayer and a feedforward neural network complex sublayer. The decoder is also composed of $N$ neural network blocks, and has a mask multiple attention layer compared with the encoder.

\begin{figure}
\centering 
\includegraphics[width=0.7\textwidth]{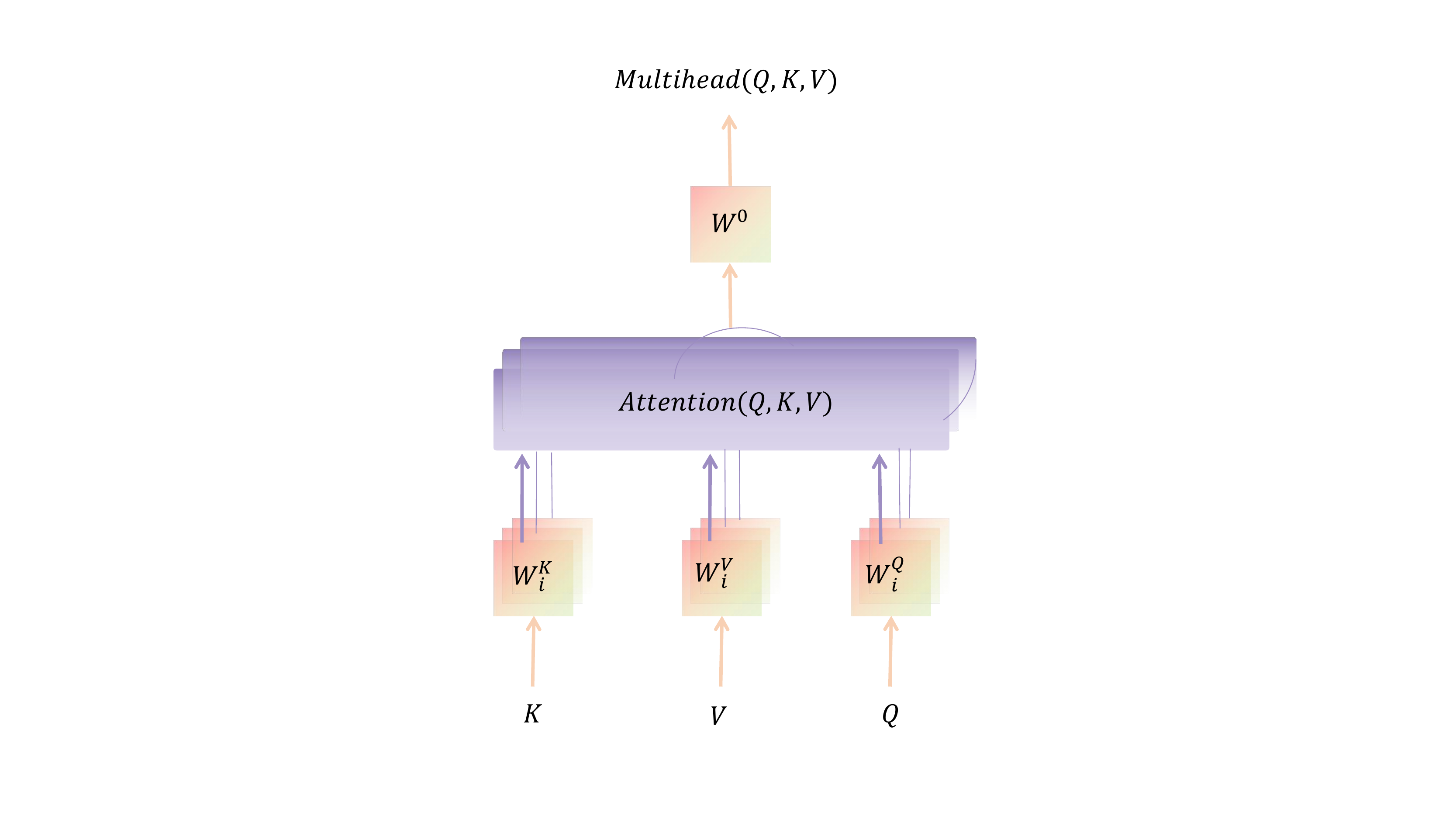}
\caption{Schematic diagram of multi-head attention mechanism} 
\label{fig:fig1}
\end{figure}

As shown in Fig. \ref{fig:fig1}, the multi-head mechanism enables the model to pay more attention to the feature information of different locations and different subspaces. The multi-head attention method consists of two steps: 1) calculation of the dot product attention; 2) multi-head attention calculation. The calculation methods of dot product attention and multiple attention are expressed as:
\begin{equation}
Attention(Q,K,V)=softmax(\frac{QK^{T}}{\sqrt{D_{k}}})V,
\end{equation}
\begin{equation}
Multihead(Q,K,V)=concat(head_{1},\dots,head_{n})W^{0},
\end{equation}

where $Q$ is the query vector, $K$ is the key vector, $V$ is the value vector, $D_{k}$ is the dimension of the hidden layer state, and $W^{0}$ is the matrix parameter. The attention value of each head is 

\begin{equation}
head_{i}=Attention(Q\cdot W_{i}^{Q},K\cdot W_{i}^{K},V\cdot W_{i}^{V}).
\end{equation}

In order to avoid the difficulty of model convergence caused by too many layers, the model adopted residual connection \cite{he2016deep} and normalized the layers (Add\&Norm) \cite{ba2016layer}. In order to better obtain the word position information, additional position encoding vectors are added to the input layer of encoder and decoder.

\subsection{Data Augmentation}
In the field of Chinese grammatical error correction, due to the limited size of the manually annotated corpus, data augmentation method is performed on the original corpus. We first counted the error types, and then used the data augmentation method \cite{Xia2022ChineseNoisyText} proposed by Tang et al. \cite{tang2021ji}, which fused word and character granularity noise, to obtain large-scale error datasets.

\subsubsection{Division of Error Types}
Specifically, we first roughly counted the error types in the open source Chinese dataset \footnote{NLPCC2018 official dataset, HSK dynamic composition corpus and SIGHAN-9 dataset} in the Chinese grammatical error correction task, and classified the Chinese syntax error types from the perspective of word and character granularity and word granularity by combining the characteristics of each error type. It can be roughly divided into 4 categories (redundancy, missing, selection, order) and 13 small categories (respectively from the word and character granularity classification, for example, redundancy is divided into word granularity and word granularity redundancy, word selection is divided into word granularity selection, word granularity close sense words, homophones and others).

\subsubsection{Data Augmentation Method from Chinese Word and Character Levels}
In order to fuse error types of different granularities, we adopted a data augmentation method that fused word and character granularity noise, and used the replication and accumulation mechanism \cite{tang2021ji} to increase the data scale. Specifically, five copies of the original data were copied, and fine-grained noise was added to the first four corpus using the above classified error types (redundancy, missing, selection and order), and comprehensive noise was added to the last corpus.

\subsection{Knowledge Distillation}
\begin{figure} 
\centering 
\includegraphics[width=0.7\textwidth]{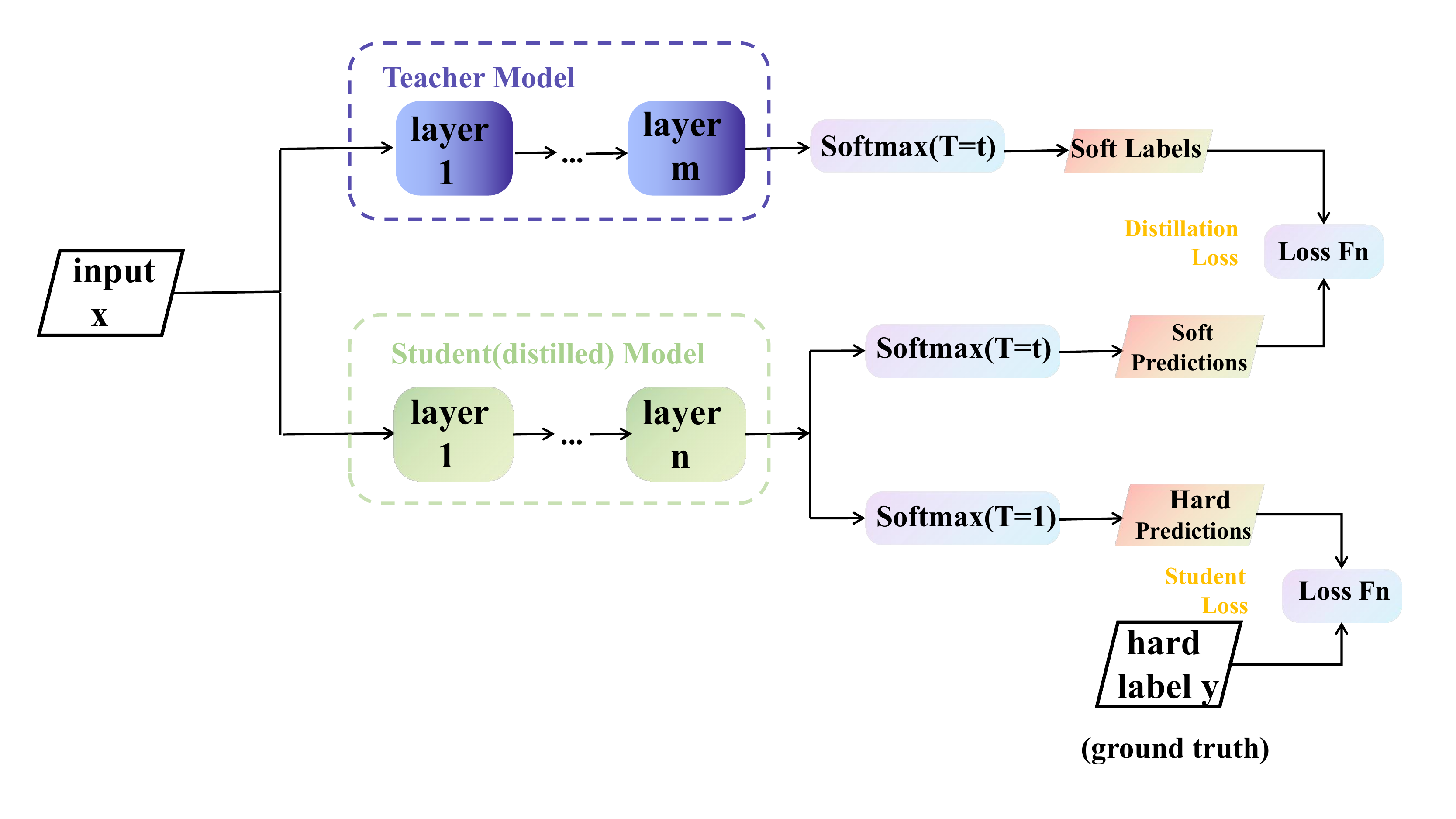}
\caption{Flow chart of knowledge distillation} 
\label{fig:fig2}
\end{figure}

Based on previous works \cite{hinton2015distilling,kim2016sequence,huang2017like}, a teacher model is used to distill a student model so that the student model can obtain the knowledge of the teacher model. The overall process is shown in Fig \ref{fig:fig2}.

Specifically, the probability distribution generated by the teacher model is taken as the soft objective, and the prediction results of the student model and the soft objective are used to calculate the loss function, which is a part of the knowledge distillation objective function. The calculation formula is as follows:

\begin{equation}\label{eq4}
    L_{soft}=\sum_{i}t_{i}\log (s_{i}), 
\end{equation}

where $t_{i}$ is the prediction result of the teacher model and $s_{i}$ is the prediction result of the student model.

Based on the setting of Hinton et al. \cite{hinton2015distilling}, we used \textit{softmax-temperature} to calculate the output probability:

\begin{equation}
    p_{i}=\frac{exp(z_{i}/T)}{\sum_{j}exp(z_{j})/T}, 
\end{equation}

where $z_{i}$ is the score of class $i$ and $T$ controls the smoothness of the output distribution. During training, $T$ applies the same temperature to student and teacher model, corresponding to soft target and soft prediction in Fig. \ref{fig:fig2}. While in reasoning, $T$ is set to 1 to recover a standard \textit{softmax}, corresponding to the prediction result in Fig. \ref{fig:fig2}.

In addition to the loss function mentioned in Eq. \ref{eq4}, the knowledge distillation objective function also includes the loss of supervised training $L_{hard}=-\sum_{i}y_{i}\log \hat{y_{i}}$, where $y_{i}$ is the target probability distribution of $i$ class samples and $\hat{y_{i}}$ is the prediction result of student model. The calculation method of knowledge distillation objective function is as follows:

\begin{equation}\label{eq6}
    L_{kd}=\alpha L_{soft}+(1-\alpha)L_{hard}, 
\end{equation}

where $L_{kd}$ is a linear combination of $L_{soft}$ and $L_{hard}$, and $\alpha$ is a hyperparameter used to adjust the relative importance of the two losses.

\subsection{Adversarial Attack}
According to the experimental results, the robustness of the current Chinese syntax error correction model is very poor. For example, when we use some adversarial samples to test the model, the performance of the model decreases a lot even though the samples are simply disturbed.

In order to test the robustness of the model, we construct the corresponding attack dataset. Inspired by the above classification of error types, we adopt redundancy, missing, and word selection as attack types, and set the proportion of attacks as 30\%. At the same time, we consider the attack of word granularity and word granularity respectively, just like the noise adding method of fusing word granularity. Specifically, first of all, random sampling the any position in each sentence, and use the Chinese synonyms tools\footnote{https://github.com/chatopera/Synonyms} to replace the position of the words, if no candidate to replace words, will be the location of words deleted. Because candidate replacement words contain both near-sense words and redundant words, we do not need to build corresponding algorithms for redundant attack types.

Examples of the three attack types are shown in Table\ref{tab:tab1}.

\begin{table}
 \caption{Attack Types}
  \centering
  \resizebox{\textwidth}{!}{
  \begin{tabular}{l|l|l}
    \toprule
    Example    & Original Text & Guan Tan Town is administratively located in Xuyi County, Jiangsu Province, China.  \\
    \midrule
    \multirow{2}{*}{Redundancy} & Word Level & Guan Tan Town is administratively \textbf{management} located in Xuyi County, Jiangsu Province, China. \\
    & Character Level & Guan Tan Town is administratively located in \textbf{in} Xuyi County, Jiangsu Province, China. \\
    \midrule
    \multirow{2}{*}{Missing} & Word Level & Guan Tan Town is administratively located Xuyi County, Jiangsu Province, China. \\
    & Character Level & Guan Tan Town is administratively located in Xuyi, Jiangsu Province, China. \\
    \midrule
    \multirow{2}{*}{Selection} & Word Level & Guan Tan Town is \textbf{city division} located in Xuyi, Jiangsu Province, China. \\
    & Character Level & Guan Tan \textbf{village} is administratively located in Xuyi County, Jiangsu Province, China. \\
    \bottomrule
  \end{tabular}}
  \label{tab:tab1}
\end{table}

\section{Experiments}
Based on the above methods, we make the following provisions: 1) we choose Chinese Wikipedia data set\footnote{https://dumps.wikimedia.org/zhwiki/} as an external corpora, and NLPCC2018\footnote{http://tcci.ccf.org.cn/conference/2018/dldoc/trainingdata02.tar.gz} official data set and HSK dynamic composition corpus \cite{zhang2006hsk} as internal corpus; 2) For all experiments, the settings of Transformer-base and Transformer-big in \cite{vaswani2017attention} are followed; 3) For the external noisy corpus, a \textit{Seq2Seq} model is trained by pre-training, and then the internal corpus is used for fine-tuning on the pre-trained model. 4) Randomly use different seeds for 5 times, and average the final score as the final result; 5) all experiments on HuggingFace Transformers library\footnote{https://github.com/huggingface/transformers} for deployment.

\subsection{Datasets and Preprocessing}
For the internal corpus, we draw 5,000 parallel corpora from 1,377,776 pairs of training statements as the development set. For the external corpus, the sentences longer than 80 words and less than 20 words were first filtered out to get 2,000,001 pairs of training statements, and 5,000 were extracted as the development set. After the replication and accumulation mechanism \cite{tang2021ji}, 10,000,005 pairs of parallel corpora are obtained and used as pre-training datasets.

\subsection{Settings}
In the Transformer-base setting, the dimension of hidden vector of model is 512 and the output dimension of the feed-forward layer is 2048, using 8 heads in the multi-head attention setting. In the Transformer- Big setting, the hidden layer size is 1024, and the output dimension of the feed-forward layer is 4096, and set the use of 16 heads in the multi-head attention. All of our models contain 6 encoder layers and 6 decoder layers.

In the distillation experiment, the teacher model is the pre-trained and fine-tuned Trans-big model, and the student model is the pre-trained Trans-base model. $\alpha$ and $T$ is set as 0.5 and 4. Dropout is set to 0.3. The optimizer adopts Adam, and the initial value of learning rate is $1\times10^{-7}$. The size of the beam search is set to 5 and the batch size is set to 64 sentences.

We take the fine-tuned Trans-big model as the teacher model, use NLPCC2018 official data and HSK dynamic composition corpus for training data, and use Eq. \ref{eq6} as the loss function of knowledge distillation to distill the Trans-base model with smaller number of parameters. It is worth noting that the Trans-base model should first go through the pre-training step.

\subsection{Metrics}
In the task of Chinese grammatical error correction, $F_{0.5}$ is usually used as the evaluation index. In this experiment, we adopt open MaxMatch (M2)\footnote{https://github.com/nusnlp/m2scorer} toolkit to calculate $F_{0.5}$, the calculation formula such as Eq. \ref{eq7} - Eq. \ref{eq9}.

\begin{equation}\label{eq7}
    P=\frac{\sum_{i=1}^{N} | e_{i} \cap ge_{i} |}{\sum_{i=1}^{N} | e_{i}|},
\end{equation}

\begin{equation}\label{eq8}
    R=\frac{\sum_{i=1}^{N} | e_{i} \cap ge_{i} |}{\sum_{i=1}^{N} | g_{i}|},
\end{equation}

\begin{equation}\label{eq9}
    F_{0.5}=\frac{(1+0.5^2)\times P \times R}{0.5^2 \times P + R},
\end{equation}

where $N$ is the total number of sentences, $e_{i}$ is the set of model modifications to the ith wrong sentence, $ge_{i}$ is the set of standard modifications to the $i$-th wrong sentence, $|e_{i} \cap ge_{i}|$ is the number of matches between model modifications to sentence $i$ and standard modifications, $P (Precision)$ represents accuracy, and $R (Recall)$ represents recall.

\subsection{Results}
The results of several experiments set in this paper are compared, as shown in Table \ref{tab:tab2}. Among them, Trans-base and Trans-big are the benchmark models, Trans-kd is the model obtained by distillation, and the architecture is consistent with the Trans-base model.

\begin{table}
 \caption{The influence of different methods on the experiment}
  \centering
  \begin{tabular}{l|llll}
    \toprule
    Model     & Parameters     & $P$  & $R$ & $F_{0.5}$\\
    \midrule
    Trans-base &141.22&	41.47&	22.62&	35.55 \\
    Trans-big&	357.83&	\textbf{42.97}&	\textbf{23.98}&	\textbf{37.10} \\
    Trans-kd&	141.22&	42.31&	23.57&	36.51 \\
    \bottomrule
  \end{tabular}
  \label{tab:tab2}
\end{table}

By comparing the performance with other NLPCC2018 Chinese grammatical error correction models (as shown in Table \ref{tab:tab3}), we find that: 1) Trans-kd model has certain competitiveness compared with trans-big; 2) The performance of trans-kd is significantly better than that of Trans-base with the same number of parameters. In addition, in view of the poor performance of the Trans-kd model compared with the Trans-fuse model on the NLPCC2018 test set, we summarized two reasons: 1) the method of source-side word Dropout \cite{tang2021ji} is missing in the distillation process; 2) Compared with the Trans-fuse model \cite{tang2021ji} set by Transformer-big \cite{vaswani2017attention}, the Trans-kd model has fewer parameters.

\begin{table}
 \caption{Performance comparison with existing models}
  \centering
  \begin{tabular}{l|lll}
    \toprule
    Model   & $P$  & $R$ & $F_{0.5}$\\
    \midrule
    Trans-kd&	42.31&	23.57&	36.51 \\
    C-Trans \cite{wang2020ji} (2020)&	38.22&	23.72&	34.05\\
    DS Trans \cite{chollampatt2018multilayer} (2020)	&39.43&	22.80&	34.41\\
    MaskGEC \cite{zhao2020maskgec} (2020)&	44.36&	22.18&	36.97 \\
    Trans-fuse \cite{tang2021ji} (2021)&	\textbf{47.29}&	\textbf{23.89}&	\textbf{39.49}\\
    \bottomrule
  \end{tabular}
  \label{tab:tab3}
\end{table}

\subsection{Ablation Study}
\subsubsection{Impact of pre-training data size on model performance}
\begin{figure}[htbp]
\centering
\subfigure[Pre-training data size versus initial $F_{0.5}$]{
	\label{fig:fig3}
	\includegraphics[scale=0.25]{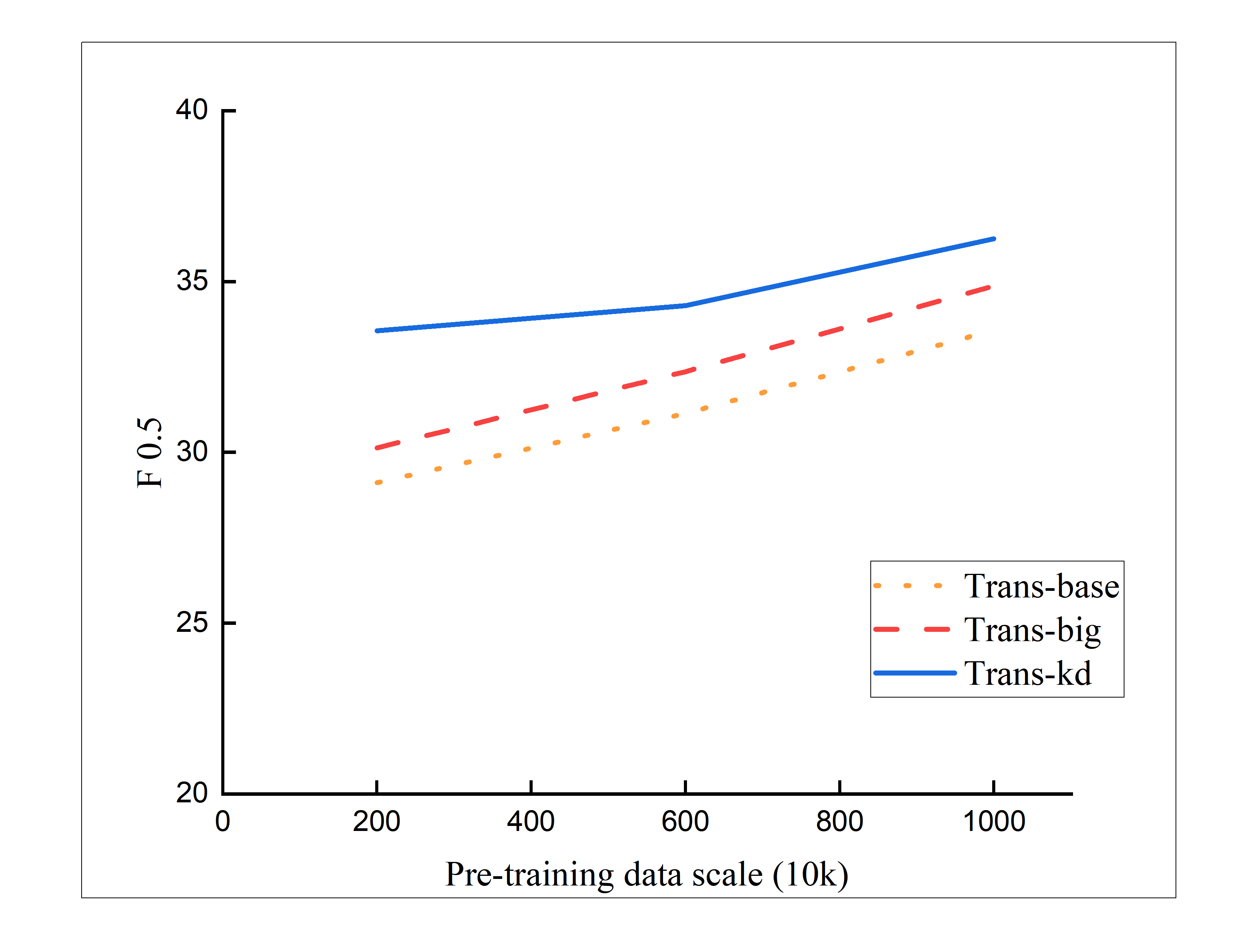}}
\subfigure[Pre-training data size versus $F_{0.5}$ after attack]{
	\label{fig:fig4}
	\includegraphics[scale=0.25]{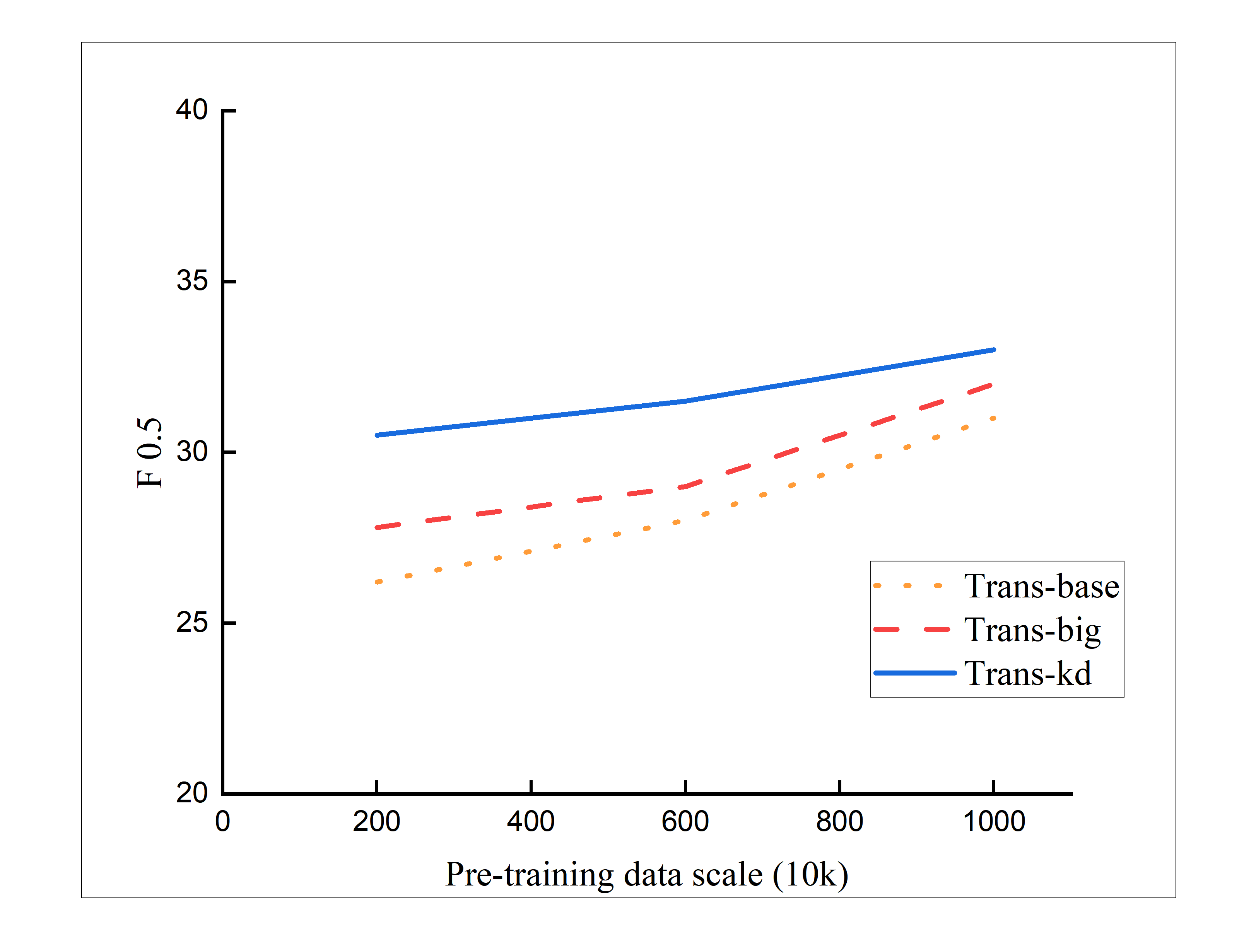}}
\caption{Curves of pre-training data size versus initial $F_{0.5}$ and $F_{0.5}$  after attack}
\label{Fig.lable}
\end{figure}

In order to analyze the influence of pre-training data size on model performance, this paper uses pre-training corpus of different sizes on standard test set and attack test set to conduct comparative experiments. Under the same parameter Settings, the experimental results are shown in Fig. \ref{fig:fig3} and Fig. \ref{fig:fig4}. The results show that the performance and robustness of the model are significantly improved with the increase of data size.

\subsubsection{Comparison of convergence speed of model training}
In order to further analyze the advantages of the distillation model, this paper analyzes the training convergence speed of the model, and takes the performance improved by the average model in each round as the evaluation index, \textit{i.e.}

\begin{equation}
    v=\frac{\Delta F_{0.5}}{\Delta epoch}.
\end{equation}

We set $\Delta epoch$ to 10, and the experimental results (as shown in Table \ref{tab:tab4}) show that the convergence rate of the Trans-kd model is significantly faster than that of Trans-base and Trans-big. Therefore, the convergence speed of Trans-kd training is significantly improved compared with the baseline model.

\begin{table}
 \caption{Comparison of convergence rate of different models}
  \centering
  \begin{tabular}{l|llll}
    \toprule
    \diagbox{Model}{Epoch}& 0$\rightarrow$ 10  & 10$\rightarrow$ 20 & 20$\rightarrow$ 30\\
    \midrule
    Trans-base &	1.61&	0.96&	0.92 \\
    Trans-big &	1.72&	0.98&	0.96 \\
    Trans-kd &	\textbf{1.76}& \textbf{1.01} &	\textbf{0.98} \\
    \bottomrule
  \end{tabular}
  \label{tab:tab4}
\end{table}

\subsubsection{Analysis of model robustness}
In order to study the influence of knowledge distillation on model robustness, we used the NLPCC2018 test set after adversarial attack under the same hyperparameter Settings, and compared the performance of Trans-base, Trans-big and Trans-kd models on these attack data sets. In addition, considering the number of model parameters, We also report the performance of the MaskGEC \cite{zhao2020maskgec} with the same number of parameters as Trans-kd on the attack dataset. The experimental results are shown in Table \ref{tab:tab5}, where refers to the difference between the post-attack test set and pre-attack test set of the model, which can show the anti-interference ability of the model. We find that the syntax error correction model with knowledge distillation has the smallest performance degradation, that is, the strongest robustness, while Trans-big and MaskGEC \cite{zhao2020maskgec}, which have better performance on the standard test set, have poor performance on the attack dataset.

\begin{table}
 \caption{The influence of different methods on the experiment}
  \centering
  \begin{tabular}{l|llll}
    \toprule
    Model     & Parameters     & Initial $F_{0.5}$ & Attack $F_{0.5} $  & $\Delta F_{0.5}$\\
    \midrule
    Trans-base&	141.22&	35.55&	32.29&	-3.26\\
    Trans-kd&	141.22&	36.51&	33.57&	\textbf{-2.94}\\
    MaskGEC \cite{zhao2020maskgec}&	141.22&	36.97&	33.52&	-3.45\\
    Trans-big&	357.83&	37.10&	33.96&	-3.14\\
    \bottomrule
  \end{tabular}
  \label{tab:tab5}
\end{table}

\section{Conclusion}
In view of the poor robustness and large scale of the existing Chinese grammatical error correction models, this paper adopts the method of knowledge distillation to solve the above problems. The output results between the student model Trans-kd and the teacher model Trans-big are constrained by the distillation loss, and the knowledge of the large model is transferred to the small model, so that the performance of the small model can be guaranteed even when the number of parameters is significantly reduced. In addition, by constraining the output results between the teacher model and the student model, the anti-attack ability of the student model can be improved. We divided the syntax error types in detail, and built the attack evaluation dataset based on the NLPCC2018 public test dataset. The experimental results show that the above method can not only reduce the number of model parameters, but also significantly improve the robustness of the model. At the same time, compared with the model with the same number of parameters, the distillation student model has faster convergence speed and better performance.


\end{document}